\documentclass[10pt,twocolumn,letterpaper]{article}

\usepackage{cvpr}
\usepackage{times}
\usepackage{epsfig}
\usepackage{graphicx}
\usepackage{amsmath}
\usepackage{amssymb}
\usepackage{algorithm}
\usepackage{algorithmic}

\usepackage[pagebackref=true,breaklinks=true,letterpaper=true,colorlinks,bookmarks=false]{hyperref}

\cvprfinalcopy 

\def\cvprPaperID{1079} 

\ifcvprfinal\fi
\begin{document}

\title{Pipeline Generative Adversarial Networks for Facial Images Generation with Multiple Attributes\vspace{-1em}}

\author{Ziqiang Zheng\quad Zhibin Yu\quad Haiyong Zheng\quad Chao Wang\quad Nan Wang\\
Ocean University of China
}
\maketitle

\begin{abstract}
Generative Adversarial Networks are proved to be efficient on various kinds of image generation tasks. However, it is still a challenge if we want to generate images precisely. Many researchers focus on how to generate images with one attribute. But image generation under multiple attributes is still a tough work. In this paper, we try to generate a variety of face images under multiple constraints using a pipeline process. The Pip-GAN (Pipeline Generative Adversarial Network) we present employs a pipeline network structure which can generate a complex facial image step by step using a neutral face image. We applied our method on two face image databases and demonstrate its ability to generate convincing novel images of unseen identities under multiple conditions previously. 
\end{abstract}
\section{Introduction}
\begin{figure}[t]
\begin{center}
\includegraphics[width=1.0\linewidth]{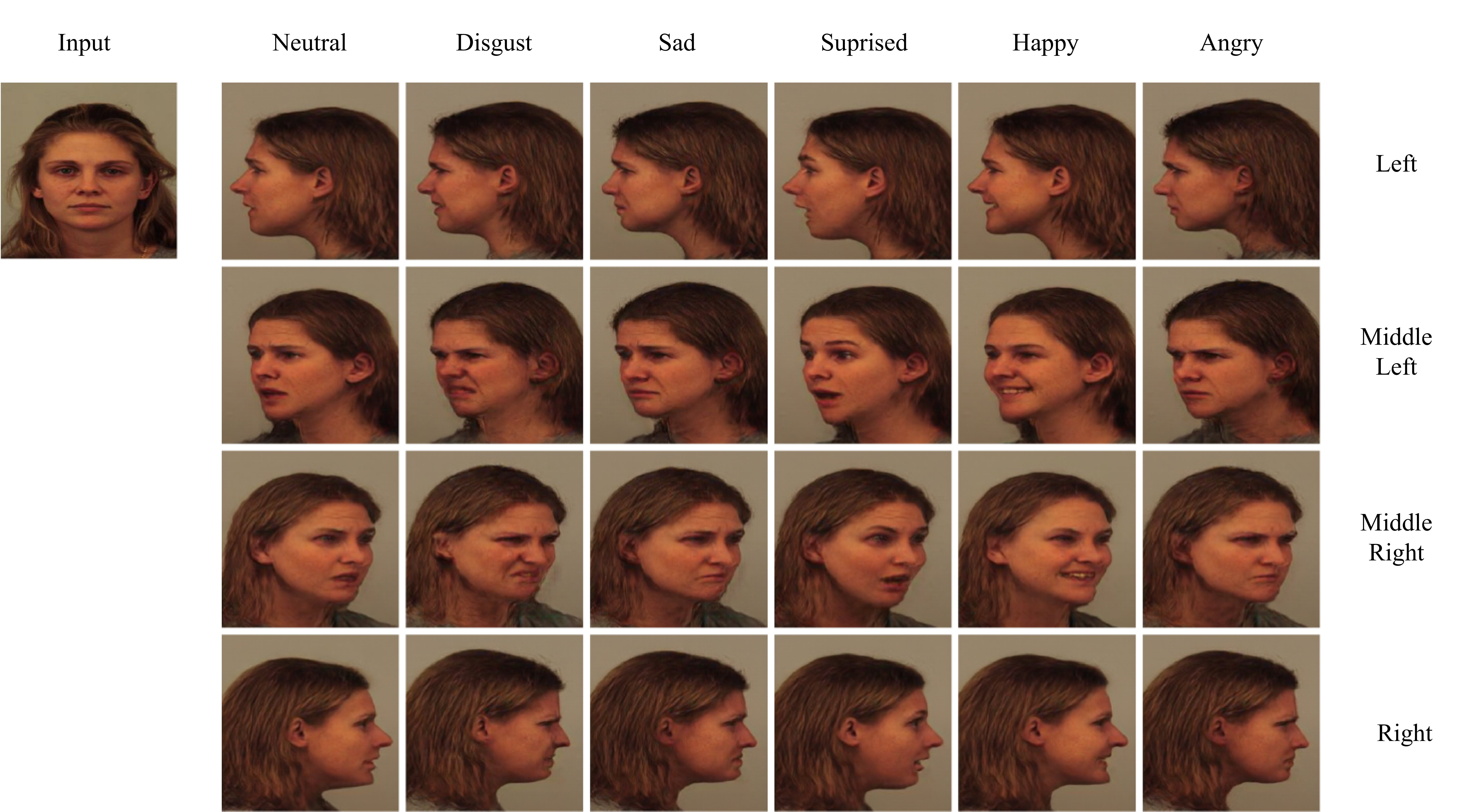}
\end{center}
\caption{The images in the first column are the neutral images and the other column images mean the different facial expression images synthesized by our model.}
\label{fig:emotionl}
\end{figure}

Since Generative Adversarial Network (GAN) was presented by Goodfellow in 2014\cite{goodfellow2014generative}, GAN is proved to be very efficient in image generation tasks in many research fields\cite{zhu2017unpaired}\cite{zhao2016energy}\cite{mirza2014conditional}\cite{press2017language}. Specially, many researchers use GAN for face images generation. Researchers show that the extension of GAN can not only generate face images with different poses\cite{yin2017towards}\cite{tran2017representation}, but also with different accessories\cite{chen2016infogan}. However, when the target images are too complex, GAN is still insufficient to achieve objects generation precisely. 
In the other hand, different facial expression face images are always required to provide samples and labels for expression recognition\cite{elfenbein2002universality}. And multi-view face images can also be used for both pose and facial expression analysis\cite{lee2006generating}\cite{calvo2008facial}. However, insufficient facial images with different pose and facial expression bring us difficulty when using supervised machine learning methods. Although human face image datasets are not uncommon, a lot of datasets only include one frontal neutral face image for each subject\cite{LFWTech}. In order to solve this problem, we aim to synthesize different facial expression images through limited neutral images\cite{antipov2017face}. Besides, image with different pose can enrich the diversity of the face samples and could be applied to the pose recognition. It is not easy to consider both pose and facial expressions simultaneously during image translation. We need a desired generative model to comprehend the current image as well as capturing the underlying data distribution. This is often a very difficult task, since a collection of image samples on one attribute may lie on a very complex manifold. Images with multiple attributes would be further more sophisticated to be modeled. As described in Figure~\ref{fig:emotionl}, the image on the left is the input image and all the right images are the target images with 2 kinds of attributes. In order to handle this task, the model should be able to generate the target images based on the input image with the appointed attributes.

\begin{figure*}
\begin{center}
\includegraphics[width=1.0\linewidth]{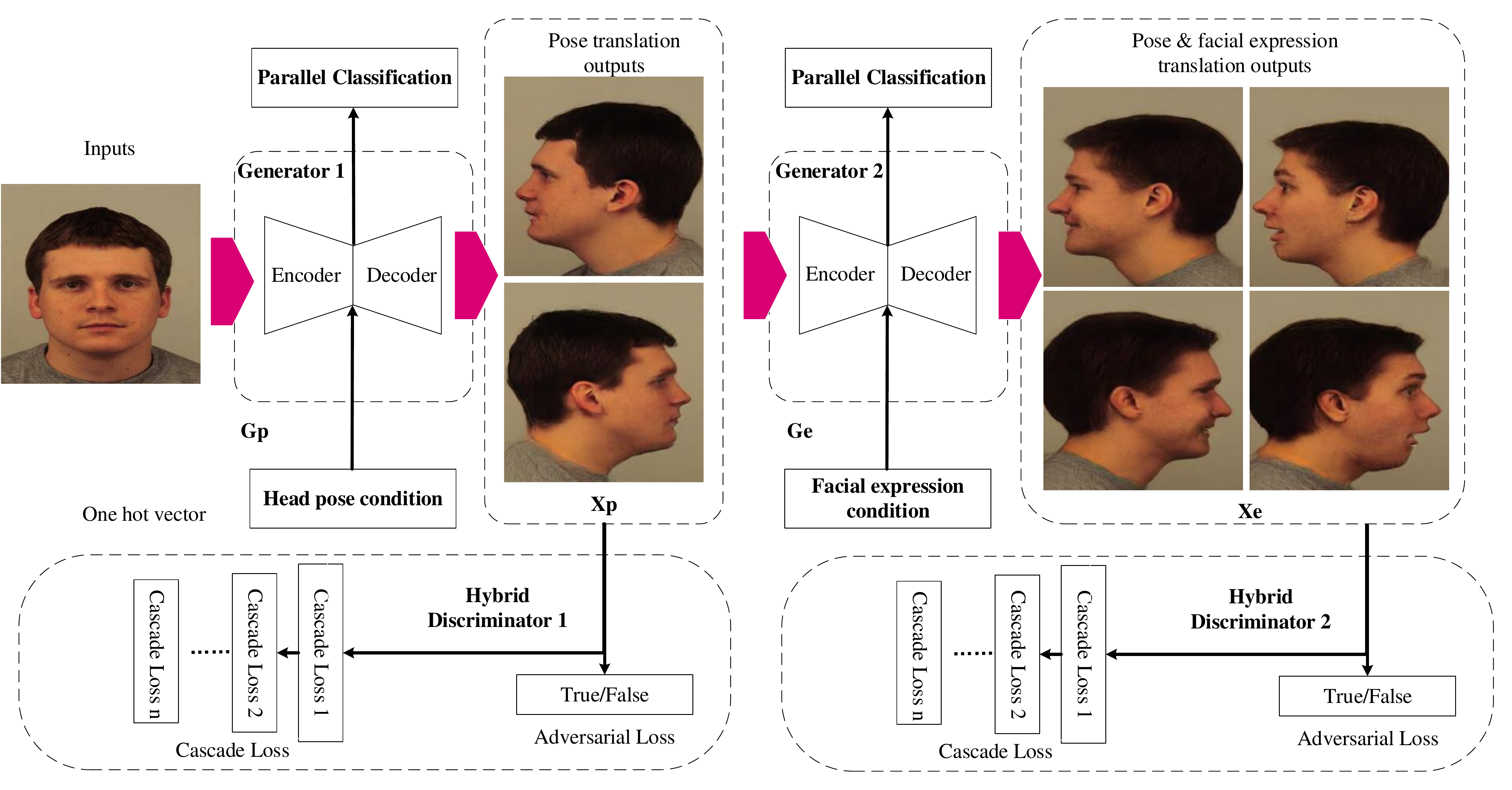}
\end{center}
\caption{The framework of our model.}
\label{fig:architecture}
\end{figure*}

Many research show that the extension of GAN is able to disentangle image information encoded by vector z and by condition y making them independent \cite{isola2016image}\cite{yi2017dualgan}. We observe that vector z is able to encode persons' facial feature such as facial pose and hair styles with traditional GAN technologies. However, the traditional GAN based image-to-image methods meet difficulty when we try to translate a facial image into an image with appointed expression and pose. 

Inspired by other image translation concepts, we proposed an end-to-end image translation framework that includes two generators and two discriminators for facial images generation with multiple attributes. These generators and discriminators make up a pipeline architecture. The detail of our work is shown in Figure.~\ref{fig:architecture}. 

The overview of our model can be found in Figure~\ref{fig:architecture}.In consideration of the difficulty of generating images with multiple attributes at the same time, we divided the synthesizing process into two stages. We assign the two different attributes: facial expression and head pose, at the two different stages respectively. We denote two image generators as the expression generator $G_{e}$ and the pose generator $G_{p}$ for the two stages separately. Each generator has its respective discriminator. We denote the expression discriminator $D_{e}$ and the pose discriminator $D_{p}$. Similar to the idea of \cite{isola2016image}, the $G_{e}$ and the $D_{e}$ is a simple image-to-image generative adversarial network and so do $G_{p}$ and $G_{e}$. And the $X_{e}$ means the expression images that are synthesized by the expression generator and so do $X_{p}$. At the first stage. We use one generators for the pose image synthesizing. The output image $X_{p}$ and real pose images are evaluated by the pose discriminator to improve the image output quality. And then we take $X_{p}$ as the input of the expression generator $G_{e}$ for the next stage. The output images from the expression generator $G_{e}$ are the final outputs of our model. Through this pipeline operation, we can break a difficult problem into several easy steps. 

It's proved that the conditional vector can represent the extension information when synthesizing images using GANs\cite{lu2017conditional}\cite{mirza2014conditional}\cite{odena2016conditional}. Zhang et al. and Reed et al.'s work also show that GANs with information are able to synthesize images conditioned on text descriptions and on spatial constraints such as bounding boxes or key points\cite{reed2016generative}\cite{zhang2016stackgan}. Following this idea, we added two conditional vectors on the coded layer of two GANs respectively. It's believed that the coded layer includes high dimensional information. By adding conditional vectors on the coded layer, we can control the facial attribute and generate the appointed synthesized images. These additional information from latent space\cite{wu2016learning} can provides the generator additional constrains and force it generates the appointed images rather than generating casually. In order to make the encoder network focus on how to select the useful features, we used an additional discriminator on the bottom neck of the encoder-decoder network for parallel classfication task to enhance the feature extraction ability of the encoder network. The detail of this conditional adversarial architecture is described in section \ref{section3.1} and \ref{section3.2}. Inspired by Chen et al.'s work\cite{chen2017photographic} and Gulrajani et al.'s work\cite{gulrajani2017improved}, we include the cascade loss and gradient penalty to improve the image generation quality \ref{section3.3} and \ref{section3.4}. The full loss function of our framework is described in section \ref{section3.5}.

Please note that there is another possible pipeline sequence of Figure~\ref{fig:architecture}. We define the sequence described in Figure~\ref{fig:architecture} as PE (pose generator at first), and we define EP (expression generator at first) as its inverse sequence structure. We evaluated these two structures in section \ref{section4}.

Our main contributions are summarized as follows:
\begin{itemize}
\item We present a new pipeline architecture that can synthesize human face images with multiple attributes. With this framework, we are able to break a tough image generation task into several easy ones.
\item We propose a parallel classification task in the bottom neck of the encoder-decoder structure to force the encoder network focus on image feature selection. 
\item We merge the cascade loss into the discriminator of our model to improve the image generation performance. An independent pre-trained cascade network is deployed in the discriminator to control the generated image quality layer by layer.
\item We combine the gradient penalty with our model for the training process to make the training more stable. 
\end{itemize}
Experimental results demonstrate the combination of these methods lead to convincing performance.

\section{Related work}
Recently, there have been remarkable progress in this field with the development of generative models that do not explicitly require this integration and can be trained using back-propagation algorithm. Two famous examples of such models are Generative Adversarial Networks (GANs)\cite{goodfellow2014generative} and Variational Autoencoders(VAE)\cite{KingmaW13}. These models are able to produce convincing image samples but not flexible enough to handle an image-to-image translation task with multiple attributes. Recent research related with GANs are mostly based on the work of DCGAN (deep convolutional generative adversarial network)\cite{radford2015unsupervised}. DCGAN has been proved to learn good feature representation from image pixels in many research\cite{zhu2017unpaired}\cite{isola2016image}. And the deep architectures have also shown the effectiveness of synthesizing photorealistic images in the adversarial networks\cite{chen2017photographic}. 

Image generation, especially face image generation, is still a hot research topic. Some early research focus on how to recover face images using incomplete face images with stacked Autoencoders and Boltzmann Machines\cite{Hinton504}. Nowadays, many researchers focus on image generation research after GANs are proposed by Goodfellow et al \cite{goodfellow2014generative}. GAN simultaneously train two models: a generative models for generating images and a discriminative model to differentiate between natural images and generated images. The key to the success of GANs is that they proposed an adversarial loss that can both optimize the generative model and discriminative model. The two-player game aims to force the generative model to synthesize more realistic images and discriminative model to differentiate between synthesized images and natural images. In the original work of Goodfellow et al. \cite{goodfellow2014generative}, GANs were used to synthesize MNIST digits and 32*32 images that aimed to reproduce the appearance of different classes in the CIFAR-10 dataset. Denton et al. \cite{denton2015deep} proposed training multiple separate GANs, one for each level in a Laplacian pyramid. Each model is trained independently to synthesize details at its scale. Assembling separately trained models in this fashion enabled the authors to synthesize smoother images and to push resolution up to 96*96. Zhao et al. \cite{mirza2014conditional} propose the conditional GANs and combine the conditional vectors with GAN to produce the images of appointed classes. In these works, researchers focus on how to convert a noise vector $z$ into an image $G(z|y)$ with a condition $y$ based on the adversarial frameworks. Reed et al.'s work further explore this field \cite{reed2016generative}. They proposed a text-to-image framework and try to convert a noise vector $z$ under a text condition $\varphi(t)$ to generate an image $G(z,\varphi(t))$ according to the text meaning. 

The recent work called InfoGAN\cite{chen2016infogan} proposed an information-theoretic extension to GANs to explore the potential of the noise vector . In their work, they decompose the input noise vector into two parts: incompressible noise $z$ and latent code $c$. And they try to make use of latent code $c$ based on the mutual information. Meanwhile, Isola et al. use a composite loss and provide an image-to-image translation solution that combines a GAN and a regression term\cite{isola2016image}. Instead of a one-dimensional conditional vector, they use a two-dimensional image as the input condition to control the input and the output performance. The authors applied their model on various datasets and demonstrate that their model is able to synthesize 256*256 images for given semantic layouts. 

Due to the development of GANs, image translation successfully drawn researchers’ attention. The key research of this topic is to find the mapping function from one image domain to another image domain\cite{kim2017learning}\cite{yi2017dualgan}, which can be used for image painting and the image style conversion. Zhu et al. proposed a method which contains two mapping functions, and relatively there are two adversarial discriminators\cite{zhu2017unpaired}. They introduced two cycle consistency losses that can capture the image domain distribution and the translation from one image domain to another image domain on unpaired image samples. Although we can collect unpaired images more easily than paired samples, we need paired samples with multiple attributes to design a precise image translation system. That's why we choose Isola et al.'s model for our research\cite{isola2016image}.

\section{Architecture\label{section3}}
Similar to Isola et al.'s work, our network includes six convolutional layers with stride-2 and a residual block. Both generator and discriminator use modules of the form convolution-BatchNorm-Relu which is mentioned in\cite{ronneberger2015u}. All Relus in the encoder are leaky.
\subsection{Conditional Adversarial Loss\label{section3.1}}
\begin{figure}[t]
\begin{center}
\includegraphics[width=1.0\linewidth]{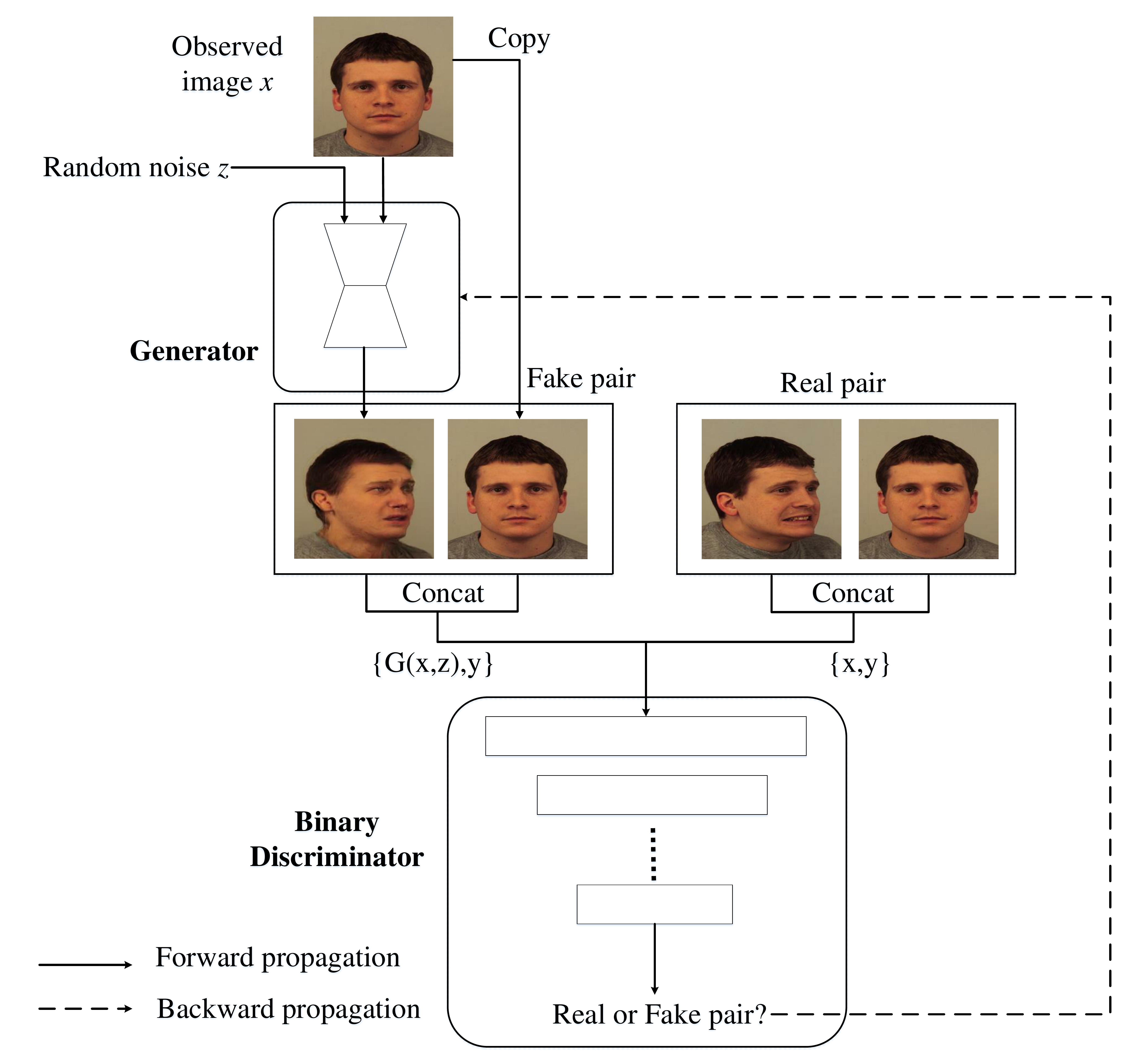}
\end{center}
\caption{The conditional adversarial loss.}
\label{fig:fig25}
\end{figure}

This objective loss with different attributes can be expressed as 
\begin{equation}
\label{eq1}
\begin{split}
L_{CGAN}(G,D)=E_{x,y\sim p_{data}(x,y)}\left[ logD\left(x,y\right)\right] \\
+ E_{x\sim p_{data}(x),z\sim p_{z}(z)}\left[log(1-D(x,G(x,z)) \right]
\end{split}
\end{equation} 
As an extension of Conditional GAN, the adversarial loss is a necessary term to our model. In Eq.~\ref{eq1}, the vector $z$ is a random noise vector following a prior noise distribution $p_{z}(z)$ and $x$ presents the observed image following the distribution $p_{data}(x)$. Here $x$ is also a 2-dimensional condition to the adversarial network. The adversarial framework try to learn a mapping function from $\{x,z\}$ to the target image $y$. The binary discriminator D try to classify between the real image pair $\{x,y\}$ and the synthesized pair $\{x, G(x,z)\}$. The detail of the conditional adversarial loss is shown in Figure.~\ref{fig:fig25}. Following Isola et al. and Mirza et al.'s work\cite{isola2016image}\cite{mirza2014conditional}, the condition $x$, which is required in both generator and discriminator sides, can force the generator to generate image $G(x,z)$ according to the observed x. The conditional advesarial loss is a basic requirement in a conditional adversarial network. We inherit this term as a part of our hybrid discriminator in our adversarial system. 

\begin{figure}[t]
\begin{center}
\includegraphics[width=1.0\linewidth]{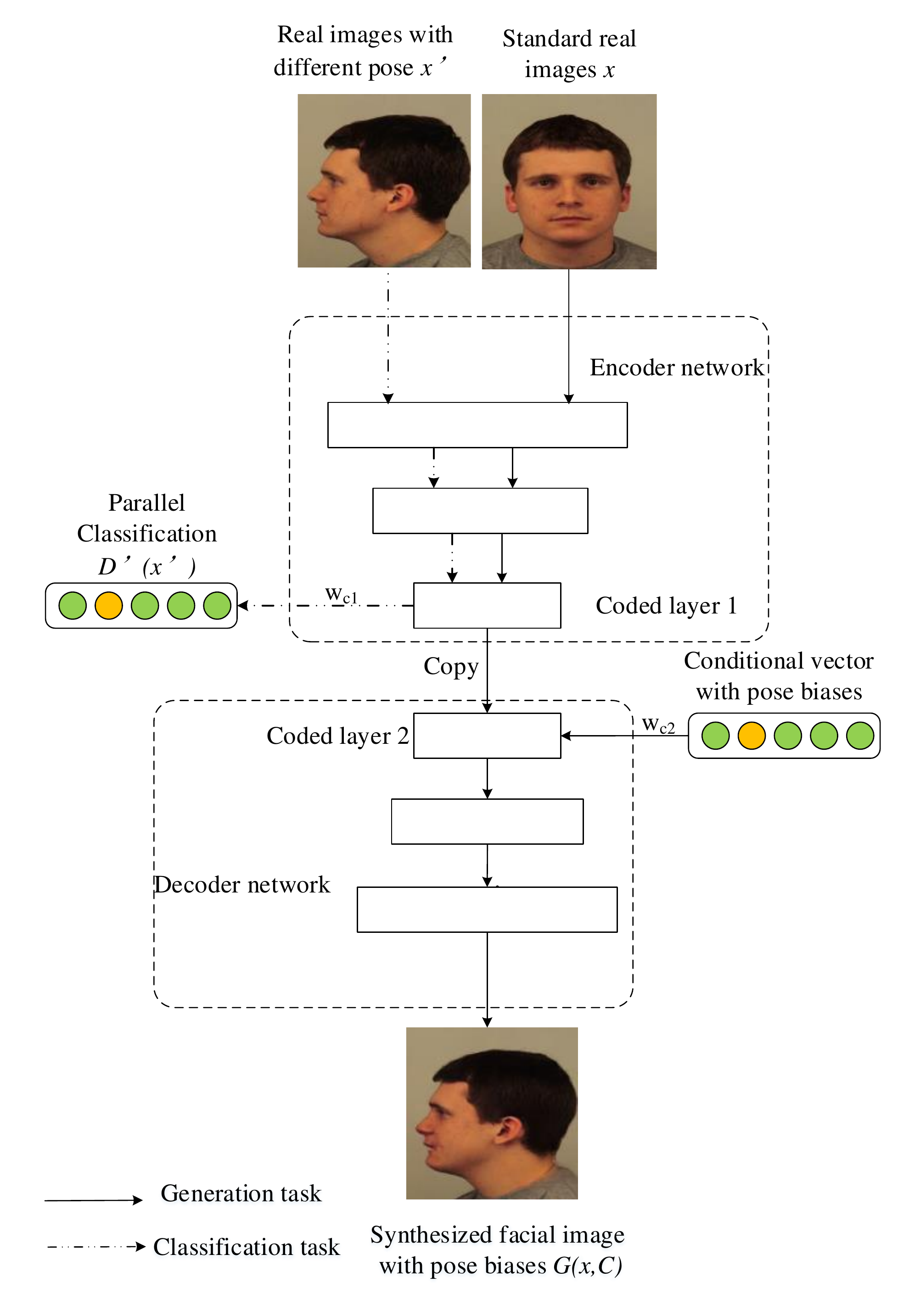}
\end{center}
\caption{The parallel classification task and the conditional biases.}
\label{fig:fig2}
\end{figure}
\subsection{Parallel Classification and Conditional Biases\label{section3.2}}
We apply a parallel classification (PC) task in the coded layer for the generator as shown in Figure. \ref{fig:fig2}. We use the cross entropy as the objective function for the classification task:
\begin{equation}
L_{PC}=\sum_{k}^{K}y_{k}log{P(y=k)}
\end{equation} 
Suppose we have $K$ classes ($K$ poses for example), the generator need to generate facial images with $K$ different poses based on the conditional biases $C$ from the coded layer 2. We introduce an additional classification task in the coded layer 1 to help the encoder network focus on how to extract the useful features from the raw images for encoding. After that, our model includes a conditional vector which indicates the appointed label we want. Please note that the conditional vector is inserted after we calculate the classification outputs and cross entropy error propagation to avoid short circuit. The pseudo code for training is listed in Algorithm \ref{algorithm1}.

\begin{algorithm}[t]
\caption{Pseudo code for parallel classification training.}
\label{algorithm1}
\begin{algorithmic}[1]
\STATE $i=p$ 
\FOR{$i=1$ to $N steps$} 
\STATE Use $x'$ to get $D'(x')$
\STATE Calculate loss for encoder network
\STATE Use $x$ and $C$ to get $G(x,C)$
\STATE Calculate loss for encoder and decoder 
\STATE Update the weights of the generator
\ENDFOR 
\end{algorithmic} 
\end{algorithm}
The classification task is implemented alternates with the image generation task. During one step, weights are updated after all the loss calcualtion of the generator are finished. 

There is a copy operation between encoder and decoder networks. The classification targets are the same as the conditional biases. But the weights $w_{c1}$ from the coded layer of the encoder network to parallel classification is different with the weights $w_{c2}$ from the conditional vector to the coded layer of the decoder network. The activation value of coded layer 1 and 2 are designed as following:
\begin{equation}
y_{c1}=f(x_{c1}+b_{c1})
\end{equation} 
\begin{equation}
y_{c2}=f(x_{c1}+w_{c2}C+b_{c2})
\end{equation} 
where $y_{c1}$ denotes the outputs of coded layer 1 and $y_{c2}$ means the outputs of coded layer 2; $x_{c1}$ is the input of coded layer 1; $b_{c1}$ and $b_{c2}$ are the layer biases; $C$ is the conditional vector which uses one-hot encoding and $f( )$ is the activation function. Please note that the conditional vector $C$ is different with the conditional image $x$ mentioned in Eq. \ref{eq1} of section \ref{section3.1}. We use $C$ to control the expression and the pose information and make our model to synthesize the facial images with appointed pose and expression class.

\subsection{Cascade Loss\label{section3.3}}

\begin{figure}[t]
\begin{center}
\includegraphics[width=1.0\linewidth]{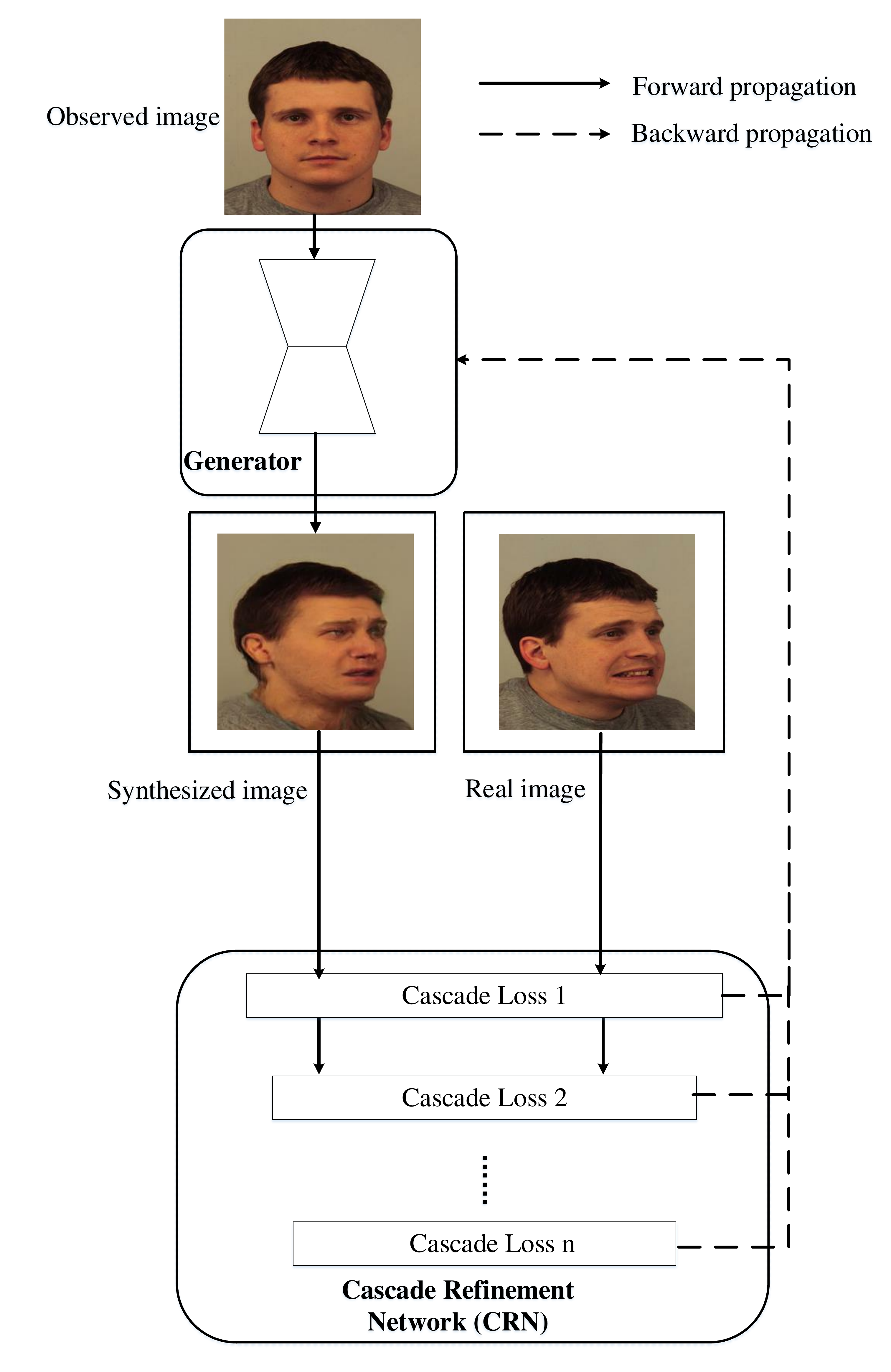}
\end{center}
\caption{The framework of cascade loss.}
\label{fig:fig3}
\end{figure}

In order to make the synthesized images clearer and perform better, we consider Chen et al.'s work and include an extra pre-trained VGG-19 network to provide the cascade loss in our hybrid discriminator\cite{chen2017photographic}. Figure.~\ref{fig:fig3} shows the detail of cascade loss. The feature maps of convolutional networks can express some important information of images. To achieve lower bound between the real data distribution and the sample distributions, the cascade refinement network can provide some important guidance for our model. The cascade loss is considered as a measurement of similarity between the target and the output images. We initialize the network with the weights and bias of the pre-trained VGG network on the Imagenet dataset\cite{deng2009imagenet}. Each layer of the network will provide cascade loss between the real target images and the synthesized images. We used the first five convolutional layers for the cascade refine network. 
\begin{equation}
\label{eq5}
{L}_{Cascade}(\theta)=\sum^{N}_{n}{\lambda}_{n}||{\Phi}_{n}(x)-{\Phi}_{n}(G((x,z);\theta))||_{1}
\end{equation}
Following the definition mentioned in Eq. \ref{eq1}, here $y$ denotes the target image and $(G(x,z))$ is the image produced by the generator $G$; $\theta$ is the parameters of the generator G; $ {\Phi}_{n}$ is the cascade response in the $n_{th}$ level in the Cascade Refinement Network (CRN). We choose the first 5 convolutional layers in VGG-19 to calculate the cascade loss. So we have $N=5$. Please note the loss ${L}_{cascade}(\theta)$ mentioned in Eq. \ref{eq5} is only used to train the parameter $\theta$ of the generator G. The CRN is a pre-trained network. The weights of CRN will not be changed during we train the generator G. The parameter ${\lambda}_{n}$ controls the influence of cascade loss in the $n_{th}$ layer of CRN.

The cascade loss can provide measure the similarity between the output and the target images under $N$ different scales and enhance the ability of discriminator. Cascade loss from the higher layer controls the global structure; the loss from the low layer controls the local detail during generation. Thus, the generator should provide better synthesized images to cheat the hybrid discriminator and finally improve the synthesized image quality. The loss constrains the generator and aims to seek the boundary between the real data distribution and sample data distribution.
\subsection{Gradient Penalty\label{section3.4}}
To further improve the performance of GANs, we consider Gulrajani et al.'s work and include the gradient penalty (GP) to our model\cite{gulrajani2017improved}. The gradient penalty can be expressed as 
\begin{equation}
L_{GP} = E_{\hat{x}\sim P_{\hat{x}}} \left[(\left \| \bigtriangledown_{\hat{x}}D(\hat{x})\right \| _{2}- 1)^{2}\right]
\end{equation}
\begin{equation}
\hat{x}=(1-\alpha)x+\alpha \tilde{x}\: \: \: \: \: where\: \alpha\sim U[0 , 1]
\end{equation}
Here $x$ represents the real sample and $\tilde{x}$ means the synthesized sample; $\alpha$ is the random parameter that controls the balance between real and fake and follows the average distribution between 0 and 1; $\bigtriangledown_{\hat{x}}$ is the gradient of $D(\hat{x})$. Please note that $\hat{x}$ is only used for calculating $L_{gp}$. We believe that the GAN with gradient penalty converge more stable and generate images with higher quality. The experiment results in section \ref{section4} also supports this point.

\subsection{Generative loss\label{section3.5}}
As mentioned in Isola et al.'s work \cite{isola2016image}, location loss is really important for synthesizing images. In our network, we use the L1 loss as the location loss. So the total loss of our generator is defined as:
\begin{equation}
\begin{split}
L = \xi_{1}L_{CGAN}(G,D) + \xi_{2}L_{Cascade} + \\
\xi_{3}L_{GP} + \xi_{4}L_{PC}+\xi_{5}\left\| y -G(x,z) \right\|_{1}
\end{split}
\end{equation}
Here $y$ is the target image and $G(x,z)$ is the image synthesized by the generator. In our experiments, we find that the model performs well when we choose $\xi_{1}=\xi_{2}=\xi_{3}=1, \xi_{4}=10, \xi_{5}=50$. And our final loss is the sum of this three different loss. We apply this final loss to the generator with Adam optimizer of learning rate $0.0002$.In our experiments. 

\section{Experimental results\label{section4}}

\subsection{Datasets}
We test our methods on three face datasets. We use two datasets to evaluate the image generation ability on pose direction and facial expression, respectively. The first dataset we used is FEI dataset\cite{thomaz2012fei}. This dataset contains 11 different pose directions from 200 individuals. Then we evaluate our methods for facial expression generation on the Yale dataset\cite{georghiades1997yale} which consist of 15 individuals and 6 different kinds of facial expression. In these two experiments, images only contains one attribute. Thus, we only use half of the model which we mentioned in Figure.~\ref{fig:architecture} for comparison and evaluation. At last, we train our model on the Karolinska Directed Emotional Faces dataset (KDEF)\cite{calvo2008facial} and evaluate the image generation ability with multiple attributes. The KDEF datasets is a set of totally 4900 pictures of human facial expressions pictures with different pose. Each subject contains 5 different pose and 7 emotional expressions (totally $5*7=35$ images). So it has 35 different face images for each person. We split the datasets into training dataset and testing dataset for 4:1.

\begin{figure}[t]
\begin{center}
\includegraphics[width=1.0\linewidth]{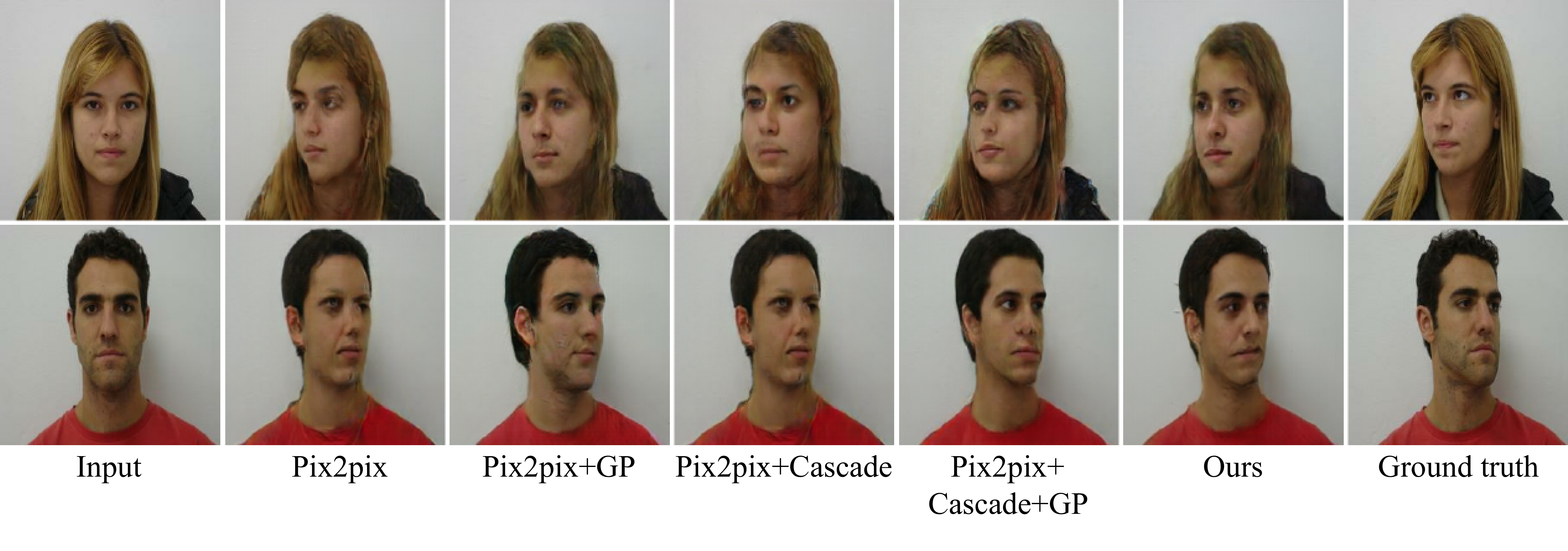}
\end{center}
\caption{Head pose generation results with different methods.}
\label{fig:fei_compare}
\end{figure}

\begin{figure*}
\begin{center}
\includegraphics[width=1.0\linewidth,height=1.0\textheight,keepaspectratio]{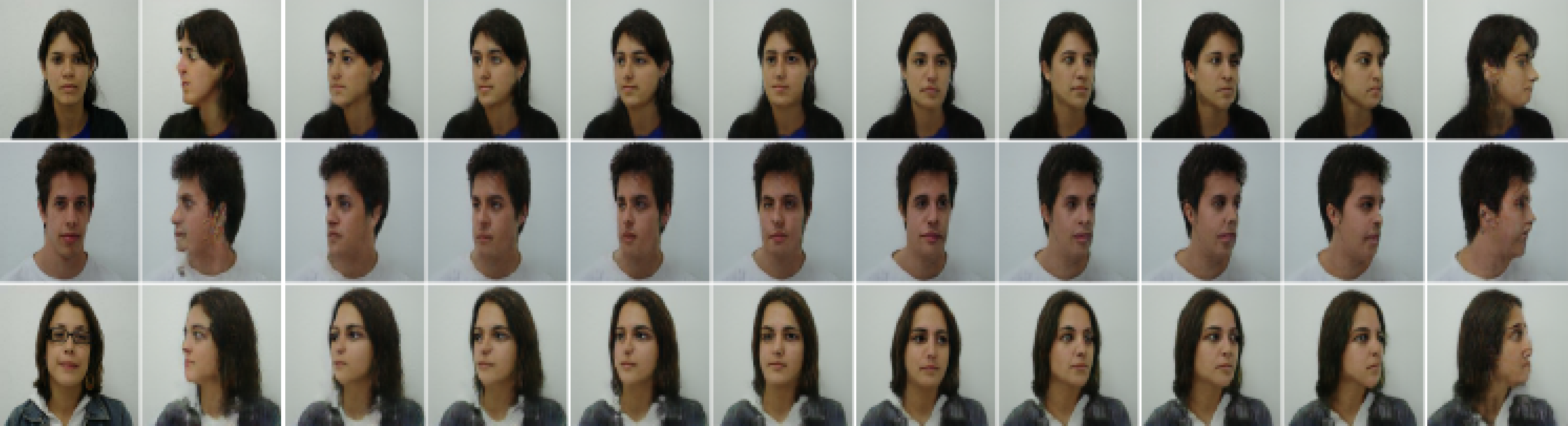}
\end{center}
\caption{Head pose generation results on 10 different angles. The images from the first column on the left are the inputs and the left images are the synthesized samples.}
\label{fig:fei_merge}
\end{figure*}

\subsection{Head pose generation\label{section4.2}}
In the FEI database, each individual contains 1 standard frontal facial image and 10 kinds of images with different pose direction. We choose the standard frontal facial image as the input for our model to generate the left 10 classes. In this task, we aim to generate images with different pose. A comparison is displayed in Figure.~\ref{fig:fei_compare}. It seems that each method is able to generate an image with the correct head direction. However, a synthesized image with correct facial detail is not an easy task. The results of Pix2pix is blur and fail to generate the eyes. The results impove when we integrate more methods mentioned in section \ref{section3}. Our methods performs better when we considered parallel classification, cascade loss and gradient penalty simultaneously.

Figure.~\ref{fig:fei_merge} shows more pose generation results on 10 different pose with our methods (Pix2pix + Cascade + GP + PC). In most of cases, our model can generate clear images with correct local facial information. The results also prove that the conditional vectors which we defined in \ref{section3.2} on the coded layer successfully controls the pose direction well.

\subsection{Facial expression generation\label{section4.3}}
\begin{figure}[t]
\begin{center}
\includegraphics[width=1.0\linewidth]{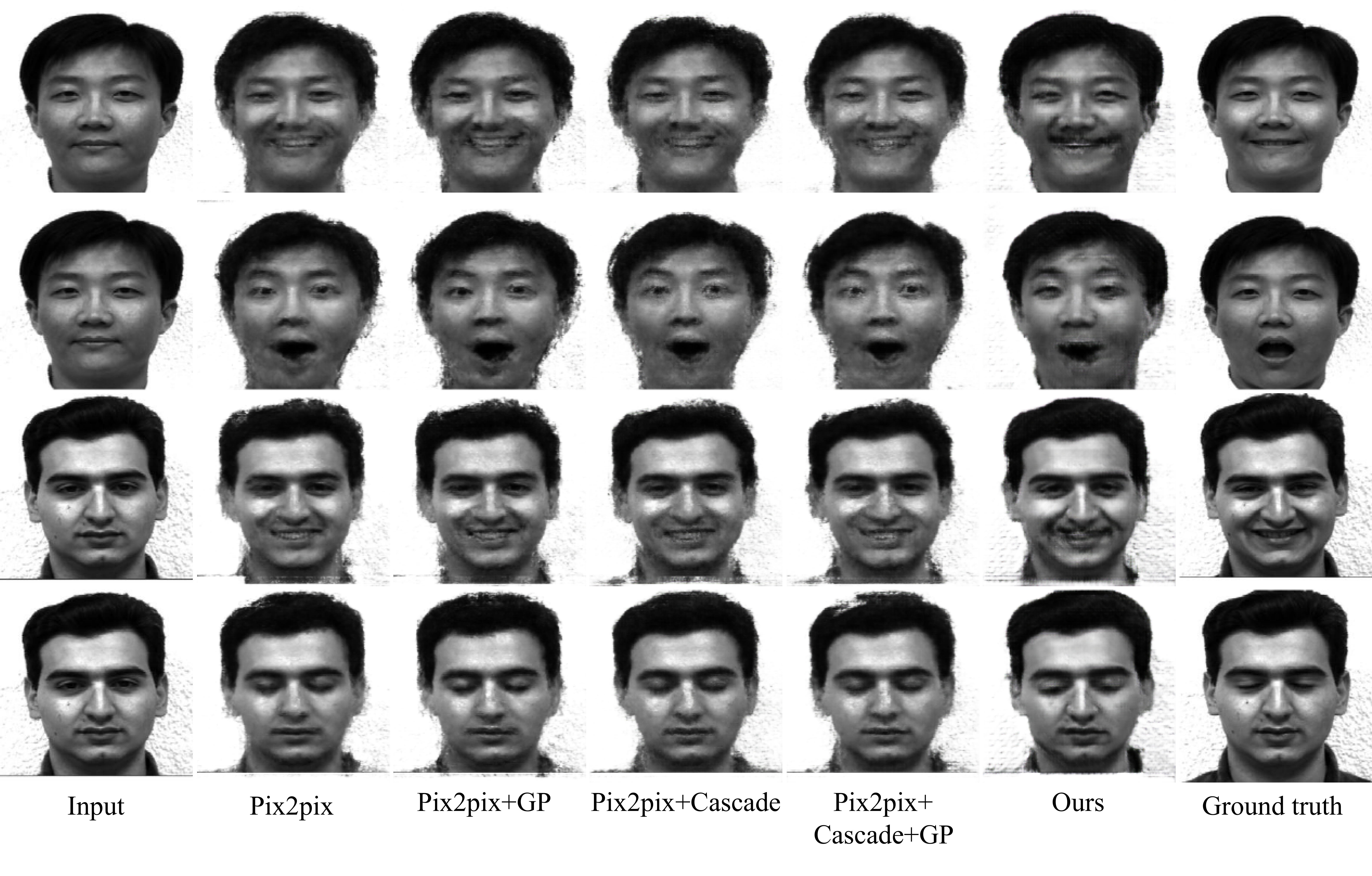}
\end{center}
\caption{Facial expression generation results with different methods.}
\label{fig:yale_compare}
\end{figure}

In the Yale dataset, we choose 1 neutral facial image as the input of our model and the left 5 different facial expression images as the targets to evaluate the expression generation performance of our model. Figure.~\ref{fig:yale_compare} shows the comparing testing results using different methods. At the first glance, each method seems to be able to generate images with the correct expression. If we enlarge the image size, we can find that the image generated by Pix2pix has a unclear boundary surrounding the subject's face. And our model can generate a clear facial image after we integrate Pix2pix with cascade loss, gradient penalty and parallel classification.

\begin{figure}[t]
\begin{center}
\includegraphics[width=1.0\linewidth]{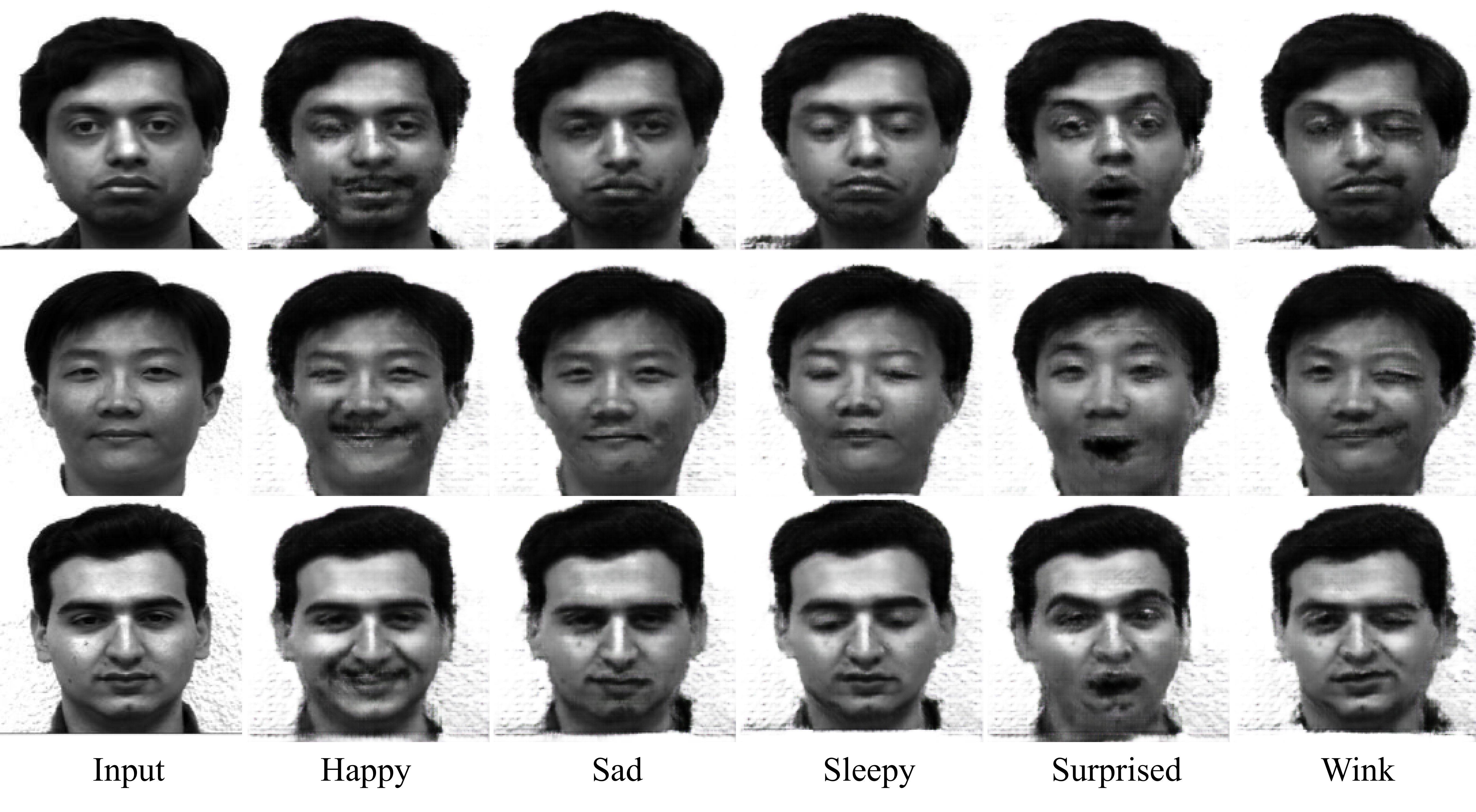}
\end{center}
\caption{Facial expression generation results on 5 different classes.}
\label{fig:yale_merge}
\end{figure}

Figure.~\ref{fig:yale_merge} shows more testing results in the Yale face datasets using our methods. These plausible generations show that our model can solve these image transfer tasks with one attribute well. 

\begin{table}
\begin{center}
\begin{tabular}{|c|c|c|c|}
\hline
Method & P-SNR &MSE &R-MSE\\
\hline\hline
Ours & 17.0296 & 0.01394& 0.1158\\
PE + Cascade +GP & 16.9893 & 0.01411 &0.1163 \\
PE + Cascade & 16.8210 & 0.01451 & 0.1183\\
PE + GP& 16.9636 & 0.01410 & 0.1166 \\
PE & 16.4461 & 0.01586 & 0.1237\\
EP+GP& 16.5461 & 0.01547 & 0.1222\\
EP& 16.4180 & 0.01557 & 0.1236\\
Pix2pix & 17.0004 & 0.01556 & 0.1230\\
\hline
\end{tabular}
\end{center}
\caption{Multiple attributes generation task using different methods on KDEF dataset.}
\label{table:comparison}
\end{table}

\begin{figure*}
\begin{center}
\includegraphics[width=1.0\linewidth,height=1.0\textheight,keepaspectratio]{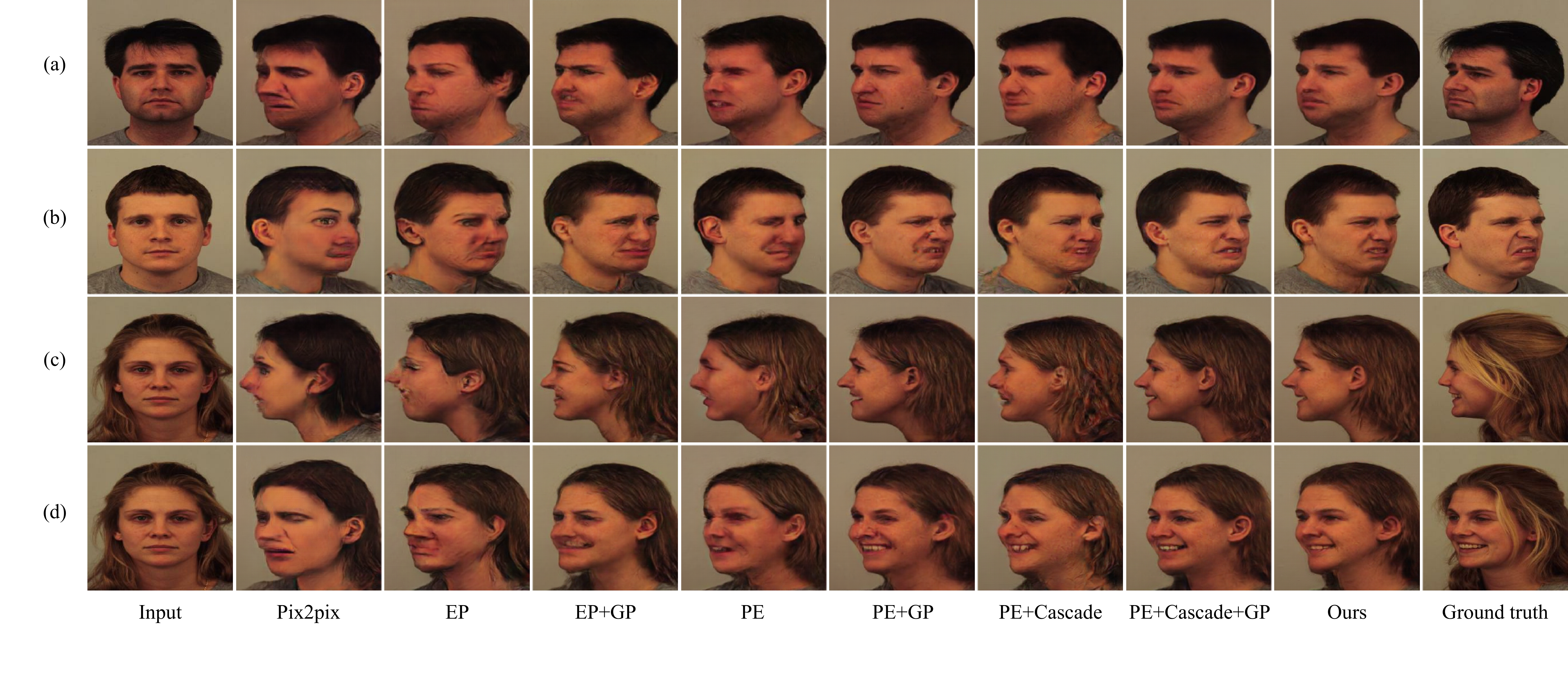}
\end{center}
\caption{Image generation results on different subjects with multiple attributes. (a) subject 1 with -45 degrees and a sad expression; (b) subject 2 with +45 degree and a disgust expression; (c) subject 3 with -90 degree and a happy expression; (d)subject 3 with -45 degree and a happy expression.}
\label{fig:fig10}
\end{figure*}

\subsection{Face generation with multiple attributes}
In this paper, our final goal is to develop a model to generate images with multiple attributes using a Pipeline generative adversarial network. We choose KDEF dataset to evaluate the performance of our model on this task. There are 4900 images including 5 different angles (-90, -45, 0, +45, +90 degrees) and 7 different expressions (afraid, angry, disgusted, happy, neutral, sad, surprised) from 70 participants (35 males and 35 females) in this dataset. Similar to the experiments mentioned in section \ref{section4.2} and \ref{section4.3}, we use one neutral straight image as the model input to generate facial images with 4 different poses using the Generator 1 (Figure.~\ref{fig:architecture}). In this step, we can generate 4 different poses with neutral expression. And these images are reused as the inputs for the Generator 2 to generate $4\time6=24$ facial images. The Table.~\ref{table:comparison} shows the image evaluation results of our experiments. Although Pix2pix has a similar P-SNR (higher is better) comparing to our methods, it has a poor performance on MSE and R-MSE (lower is better on both criteria). Our model, which integrate PE structure, Cascade loss, gradient penalty and parallel classification, has the best performs when considering cascade loss, gradient penalty, cross entropy and Pose-Emotion (PE) pipeline structure simultaneously. It should be noted that we find that PE structure performs better than EP structure. By using this trick, our results can be clearer and can generate elaborate images with plausible parts such as teeth, eyes and so on. Because we find that pose generation task on KDEF has better performance than facial expression generation task on the Generator 1. The generation quality of the first generator will influent the performance of the next generator. That's why we choose PE structure in Figure.~\ref{fig:architecture}.

We evaluate the generation performance with multiple attribution of each model in Figure.~\ref{fig:fig10}. The first column image is the input of each model, other images represent outputs. (a), (b) and (c) show the performance of each model using 3 different subjects. (c) and (d) represent the generation results using the same subject with different attribute outputs (-90 degrees + happy). In the second column, we can find that the results of a single Pix2pix model are blur with distorted facial features. In this task, we try to generate $4\time6=24$ kinds of images with a single Pix2pix model. The result show that it's possible for a Pix2pix model to generate a facial image with correct head pose. But it fail to generate the correct expression in (b), (c) and (d) cases. Images in the third and the fifth column show the results of a pipeline Pix2pix structure of PE (pose first) and EP (expression first), respectively. In these two tasks, one Pix2pix model is only used to generate images with one attribute (pose or expression). The results of PE and EP perform better than a single Pix2pix case. We guess that the expression image generation is more difficult than pose image generation on KDEF dataset. However, the image detail of these two models are still very poor. Relatively, PE performs better than EP structure. The mouth generated by PE model is more approaching to the ground truth. We get better image quality when we include gradient penalty on PE and EP structure. We further enhanced the generation quality when we integrate GP and cascade loss on PE structure. And our method achieves the best performance among these methods.

\section{Conclusion}
In this paper, we proposed a new architecture that can generate facial images with multiple different attributes. And the pipeline architecture with two different generators and two discriminators can reduce the difficulty of each generator and guarantee the final image translation results. This direct approach can divide the difficult task into two stages and perform better than contemporaneous work single GAN based model. Furthermore, we successfully improve the image generation performance by adding cascade loss, gradient penalty and parallel classification. Our method also provide a new way for image data augmentation under multiple attributes.


Nowadays, unsupervised learning is very popular because its good performance in representation learning \cite{zhu2017unpaired}. On the other side, summarizing multiple attributes from images is not an easy work if the image is complex. How to collect attributes automatically through unsupervised learning is an interesting work. We leave this part in our future work.

{\small
\bibliographystyle{ieee}
\bibliography{egbib}

\begin{thebibliography}{10}\itemsep=-1pt

\bibitem{antipov2017face}
G.~Antipov, M.~Baccouche, and J.-L. Dugelay.
\newblock Face aging with conditional generative adversarial networks.
\newblock {\em arXiv preprint arXiv:1702.01983}, 2017.

\bibitem{calvo2008facial}
M.~G. Calvo and D.~Lundqvist.
\newblock Facial expressions of emotion (kdef): Identification under different
  display-duration conditions.
\newblock {\em Behavior research methods}, 40(1):109--115, 2008.

\bibitem{chen2017photographic}
Q.~Chen and V.~Koltun.
\newblock Photographic image synthesis with cascaded refinement networks.
\newblock {\em arXiv preprint arXiv:1707.09405}, 2017.

\bibitem{chen2016infogan}
X.~Chen, Y.~Duan, R.~Houthooft, J.~Schulman, I.~Sutskever, and P.~Abbeel.
\newblock Infogan: Interpretable representation learning by information
  maximizing generative adversarial nets.
\newblock In {\em Advances in Neural Information Processing Systems}, pages
  2172--2180, 2016.

\bibitem{deng2009imagenet}
J.~Deng, W.~Dong, R.~Socher, L.-J. Li, K.~Li, and L.~Fei-Fei.
\newblock Imagenet: A large-scale hierarchical image database.
\newblock In {\em Computer Vision and Pattern Recognition, 2009. CVPR 2009.
  IEEE Conference on}, pages 248--255. IEEE, 2009.

\bibitem{denton2015deep}
E.~L. Denton, S.~Chintala, R.~Fergus, et~al.
\newblock Deep generative image models using a￼ laplacian pyramid of
  adversarial networks.
\newblock In {\em Advances in neural information processing systems}, pages
  1486--1494, 2015.

\bibitem{elfenbein2002universality}
H.~A. Elfenbein and N.~Ambady.
\newblock On the universality and cultural specificity of emotion recognition:
  a meta-analysis.
\newblock {\em Psychological bulletin}, 128(2):203, 2002.

\bibitem{georghiades1997yale}
A.~Georghiades, P.~Belhumeur, and D.~Kriegman.
\newblock Yale face database.
\newblock {\em Center for computational Vision and Control at Yale University,
  http://cvc. yale. edu/projects/yalefaces/yalefa}, 2, 1997.

\bibitem{goodfellow2014generative}
I.~Goodfellow, J.~Pouget-Abadie, M.~Mirza, B.~Xu, D.~Warde-Farley, S.~Ozair,
  A.~Courville, and Y.~Bengio.
\newblock Generative adversarial nets.
\newblock In {\em Advances in neural information processing systems}, pages
  2672--2680, 2014.

\bibitem{gulrajani2017improved}
I.~Gulrajani, F.~Ahmed, M.~Arjovsky, V.~Dumoulin, and A.~Courville.
\newblock Improved training of wasserstein gans.
\newblock {\em arXiv preprint arXiv:1704.00028}, 2017.

\bibitem{Hinton504}
G.~E. Hinton and R.~R. Salakhutdinov.
\newblock Reducing the dimensionality of data with neural networks.
\newblock {\em Science}, 313(5786):504--507, 2006.

\bibitem{LFWTech}
G.~B. Huang, M.~Ramesh, T.~Berg, and E.~Learned-Miller.
\newblock Labeled faces in the wild: A database for studying face recognition
  in unconstrained environments.
\newblock Technical Report 07-49, University of Massachusetts, Amherst, October
  2007.

\bibitem{isola2016image}
P.~Isola, J.-Y. Zhu, T.~Zhou, and A.~A. Efros.
\newblock Image-to-image translation with conditional adversarial networks.
\newblock {\em arXiv preprint arXiv:1611.07004}, 2016.

\bibitem{kim2017learning}
T.~Kim, M.~Cha, H.~Kim, J.~Lee, and J.~Kim.
\newblock Learning to discover cross-domain relations with generative
  adversarial networks.
\newblock {\em arXiv preprint arXiv:1703.05192}, 2017.

\bibitem{KingmaW13}
D.~P. Kingma and M.~Welling.
\newblock Auto-encoding variational bayes.
\newblock {\em CoRR}, abs/1312.6114, 2013.

\bibitem{lee2006generating}
H.-S. Lee and D.~Kim.
\newblock Generating frontal view face image for pose invariant face
  recognition.
\newblock {\em Pattern Recognition Letters}, 27(7):747--754, 2006.

\bibitem{lu2017conditional}
Y.~Lu, Y.-W. Tai, and C.-K. Tang.
\newblock Conditional cyclegan for attribute guided face image generation.
\newblock {\em arXiv preprint arXiv:1705.09966}, 2017.

\bibitem{mirza2014conditional}
M.~Mirza and S.~Osindero.
\newblock Conditional generative adversarial nets.
\newblock {\em arXiv preprint arXiv:1411.1784}, 2014.

\bibitem{odena2016conditional}
A.~Odena, C.~Olah, and J.~Shlens.
\newblock Conditional image synthesis with auxiliary classifier gans.
\newblock {\em arXiv preprint arXiv:1610.09585}, 2016.

\bibitem{press2017language}
O.~Press, A.~Bar, B.~Bogin, J.~Berant, and L.~Wolf.
\newblock Language generation with recurrent generative adversarial networks
  without pre-training.
\newblock {\em arXiv preprint arXiv:1706.01399}, 2017.

\bibitem{radford2015unsupervised}
A.~Radford, L.~Metz, and S.~Chintala.
\newblock Unsupervised representation learning with deep convolutional
  generative adversarial networks.
\newblock {\em arXiv preprint arXiv:1511.06434}, 2015.

\bibitem{reed2016generative}
S.~Reed, Z.~Akata, X.~Yan, L.~Logeswaran, B.~Schiele, and H.~Lee.
\newblock Generative adversarial text to image synthesis.
\newblock {\em arXiv preprint arXiv:1605.05396}, 2016.

\bibitem{ronneberger2015u}
O.~Ronneberger, P.~Fischer, and T.~Brox.
\newblock U-net: Convolutional networks for biomedical image segmentation.
\newblock In {\em International Conference on Medical Image Computing and
  Computer-Assisted Intervention}, pages 234--241. Springer, 2015.

\bibitem{thomaz2012fei}
C.~E. Thomaz.
\newblock Fei face database.
\newblock {\em online] http://fei. edu. br/\~{} cet/facedatabase. html
  (accessed 2 October 2012)}, 2012.

\bibitem{tran2017representation}
L.~Tran, X.~Yin, and X.~Liu.
\newblock Representation learning by rotating your faces.
\newblock {\em arXiv preprint arXiv:1705.11136}, 2017.

\bibitem{wu2016learning}
J.~Wu, C.~Zhang, T.~Xue, B.~Freeman, and J.~Tenenbaum.
\newblock Learning a probabilistic latent space of object shapes via 3d
  generative-adversarial modeling.
\newblock In {\em Advances in Neural Information Processing Systems}, pages
  82--90, 2016.

\bibitem{yi2017dualgan}
Z.~Yi, H.~Zhang, P.~T. Gong, et~al.
\newblock Dualgan: Unsupervised dual learning for image-to-image translation.
\newblock {\em arXiv preprint arXiv:1704.02510}, 2017.

\bibitem{yin2017towards}
X.~Yin, X.~Yu, K.~Sohn, X.~Liu, and M.~Chandraker.
\newblock Towards large-pose face frontalization in the wild.
\newblock {\em arXiv preprint arXiv:1704.06244}, 2017.

\bibitem{zhang2016stackgan}
H.~Zhang, T.~Xu, H.~Li, S.~Zhang, X.~Huang, X.~Wang, and D.~Metaxas.
\newblock Stackgan: Text to photo-realistic image synthesis with stacked
  generative adversarial networks.
\newblock {\em arXiv preprint arXiv:1612.03242}, 2016.

\bibitem{zhao2016energy}
J.~Zhao, M.~Mathieu, and Y.~LeCun.
\newblock Energy-based generative adversarial network.
\newblock {\em arXiv preprint arXiv:1609.03126}, 2016.

\bibitem{zhu2017unpaired}
J.-Y. Zhu, T.~Park, P.~Isola, and A.~A. Efros.
\newblock Unpaired image-to-image translation using cycle-consistent
  adversarial networks.
\newblock {\em arXiv preprint arXiv:1703.10593}, 2017.

\end{thebibliography}
}

\end{document}